\documentclass{article} 
\usepackage{nips13submit_e,times}

\usepackage{amsfonts}
\usepackage{amsmath}
\usepackage{bm}
\usepackage{graphicx}
\usepackage{hyperref}
\usepackage{natbib}
\usepackage{url}

\usepackage{caption}
\usepackage{subcaption}

\usepackage{algorithm}
\usepackage{algorithmic}

\DeclareMathOperator{\tr}{tr}
\DeclareMathOperator*{\argmax}{arg\,max}

\title{Bayesian Inference in Sparse Gaussian Graphical Models}

\author{
Peter Orchard \\
University of Edinburgh \\
\texttt{P.R.Orchard@sms.ed.ac.uk} \\
\And
Felix Agakov \\
Pharmatics Ltd, Edinburgh, UK \\
\texttt{felix@pharmaticsltd.com} \\
\And
Amos Storkey \\
University of Edinburgh \\
\texttt{a.storkey@ed.ac.uk} \\
}

\nipsfinalcopy

\begin{document}

\maketitle

\begin{abstract}
One of the fundamental tasks of science is to find explainable relationships
between observed phenomena. One approach to this task that has received
attention in recent years is based on probabilistic graphical modelling with
sparsity constraints on model structures. In this paper, we describe two new
approaches to Bayesian inference of sparse structures of Gaussian graphical
models (GGMs). One is based on a simple modification of the cutting-edge block
Gibbs sampler for sparse GGMs, which results in significant computational gains
in high dimensions. The other method is based on a specific construction of the
Hamiltonian Monte Carlo sampler, which results in further significant
improvements. We compare our fully Bayesian approaches with the popular
regularisation-based graphical LASSO, and demonstrate significant advantages of
the Bayesian treatment under the same computing costs. We apply the methods to
a broad range of simulated data sets, and a real-life financial data set.
\end{abstract}

\section{Introduction}

Learning and exploiting structure in data is a fundamental machine learning
problem. The notion of structure is closely related to sparsity. Indeed, to
find structure often means to discover a sparse underlying graphical model.
Learning a model structure, and incorporating prior structural knowledge, is
important for prediction, knowledge discovery, and computational tractability,
and has applications across multiple domains such as finance and systems
biology.

A popular technique for learning sparse graphical models is to optimise an
objective function with a penalty on the $L1$ norm of the parameters. This has
been applied, for example, to regression \citep{tibshirani_1996_regression},
and to sparse coding \citep{olshausen_1996_emergence}. There has been
considerable recent attention on Gaussian graphical models (GGMs), wherein the
inverse covariance (precision) matrix reflects the structure: zeros in the
precision correspond to missing edges in the model's Markov random field.
Therefore, the problem of structure learning in a GGM may be cast as the
learning of a sparse precision matrix. Much recent work has been devoted to the
MAP solution of an $L1$-constrained formulation of this problem
\citep{meinshausen_2006_high,banerjee_2008_model,friedman_2008_sparse,
duchi_2008_projected,scheinberg_2010_sparse}; and its extensions to group
sparsity \citep{friedman_2010_applications}; handling non-Gaussianity
\citep{liu_2009_nonparanormal}; and incorporating latent variables
\citep{agakov_2012_discriminative}.

Bayesian methods may offer a more principled approach, but they are often
considered to be much slower than optimisation methods. However,
\citet{mohamed_2012_bayesian} recently compared the $L1$ approach with Bayesian
methods based on the ``spike-and-slab'' prior, focussing on unsupervised linear
latent variable models. They found that the Bayesian methods could outperform
$L1$, even when both were constrained by the same time budget. Here, we address
the question of whether Bayesian methods can outperform the $L1$ optimisation
approach to infer sparse precision matrices.

A class of Bayesian GGMs based on the GWishart distribution (see section
\ref{sec:gw}) has been under development in parallel with the graphical lasso.
See, for example,
\citep{roverato_2002_hyper,atay-kayis_2005_monte,wang_2010_simulation,
dobra_2010_modeling,mitsakakis_2011_metropolis,lenkoski_2011_computational,
dobra_2011_bayesian, wang_2012_efficient}. Inference in these models is often
limited by the efficiency with which the GWishart can be sampled.
\citet{wang_2012_efficient} demonstrate that the block Gibbs sampler is
currently state of the art for this task. A more efficient sampler would make
inference in GWishart-based models faster, and could make practical the use of
more complex, higher-dimensional models. Hamiltonian Monte Carlo (HMC) samplers
(see section \ref{sec:hmc}) can facilitate fast mixing in distributions where
the random variables are strongly coupled. Furthermore, they naturally take
advantage of sparsity: the bottleneck in HMC is often the computation of the
energy gradient with respect to the distribution parameters. Fewer parameters
means fewer gradients to evaluate. In this paper, we develop an HMC approach to
sample from the GWishart distribution.

The key contributions of the paper are as follows. First, we identify a simple 
modification of the block Gibbs sampler that makes it considerably more
efficient when used as part of a high-dimensional GGM sampler. We then develop
an HMC approach to sampling the GWishart, and demonstrate substantial gains in
efficiency over block Gibbs methods. Finally, we utilise this sampler within a
sparse Bayesian GGM, and show on a real-world data set that the model can
outperform graphical lasso, even when the methods have the same time budget.

\section{Background}

\subsection{Graphical Lasso}

The inverse covariance (precision) matrix $\bm{\Lambda}$ of a Gaussian model
$p(\bm{y}) = \mathcal{N}(\bm{y}|\bm{\mu},\bm{\Lambda}^{-1})$ determines its
graphical structure: zeros in $\bm{\Lambda}$ correspond to missing edges in the
Markov random field. Therefore one may hope to learn the model structure by
learning a sparse precision. One way to do this is to introduce $L1$ penalties
on the elements of $\bm{\Lambda}$, and optimise the objective
\begin{equation}
 \argmax_{\bm{\Lambda}} \left[
  \log\det\bm{\Lambda}
  - \tr\left( \bm{S} \bm{\Lambda} \right)
  - \gamma \| \bm{\Lambda} \|_1 \right] ,
\end{equation}
where $\bm{S}$ is the empirical covariance, $\gamma$ is the penalty, and
$\| \bm{\Lambda} \|_1 = \sum_{i,j} |\Lambda_{ij}|$ is the $L1$ norm of
$\bm{\Lambda}$. The optimum is the \textit{maximum a posteriori} (MAP)
solution for $\bm{\Lambda}$ with independent Laplace priors on its elements.
The penalty $\gamma$ must be fixed in advance. Typically, this is done by
cross-validation over some plausible range.

There has been much recent work on solving this problem efficiently; see, for
example,
\citep{meinshausen_2006_high,banerjee_2008_model,friedman_2008_sparse,
duchi_2008_projected,scheinberg_2010_sparse}. The approach of
\citet{friedman_2008_sparse} was named the graphical lasso, and the problem
itself is often referred to by the same name - as we do in this paper.

\subsection{The GWishart Distribution}
\label{sec:gw}

Let $G = (V,E)$ denote a graph. The GWishart $W_G(b,\bm{D})$ is a distribution
over matrices $\bm{\Lambda}$ that respect the graph structure. Precisely, the
density is:
\begin{equation}
 p(\bm{\Lambda}|\bm{G}) = \frac{1}{I_G(b,\bm{D})}
  |\bm{\Lambda}|^{\frac{b-2}{2}}
  \exp\left[ -\frac{1}{2}\tr(\bm{D}\bm{\Lambda}) \right]
  1_{  [\bm{\Lambda} \in M^{+}(G)] } \ ,
\end{equation}
where $b$ is the degrees of freedom, $\bm{D}$ is the scale matrix, $1_{[]}$
is the indicator function,
$M^{+}(G) = \{ \bm{\Lambda} \in S_{++}
 \mbox{ such that } \Lambda_{ij} = 0 \mbox{ if } (i,j) \notin E, i \neq j \}$,
and $S_{++}$ is the cone of positive-definite matrices. Computing the
normalisation constant $I_G(b,\bm{D})$ is not straightforward. It
may be approximated by sampling \citep{atay-kayis_2005_monte} or by Laplace
approximation \citep{lenkoski_2011_computational}, for example.

Following the notation of \citet{atay-kayis_2005_monte}, we define
\begin{align}
 \mathcal{V} &= \{ (i,j), i \leq j
  \ \mbox{such that either}\ i=j, i \in V \ \mbox{or} \ (i,j) \in E \} , \\
 \mathcal{W} &= \{ (i,j), i,j \in V, i \leq j \} , \\
 \bar{\mathcal{V}} &= \mathcal{W} \setminus \mathcal{V} .
\end{align}
We define
$\bm{\Lambda}^{\mathcal{V}} = \{ \Lambda_{ij} : (i,j) \in \mathcal{V} \}$.
We use $\bm{\Lambda}_{\mathcal{V}}$ to refer to a vectorised form of
$\bm{\Lambda}^{\mathcal{V}}$.

Like the Wishart, the GWishart is the conjugate prior of the precision of a
multivariate Gaussian. Let $\bm{y} \sim \mathcal{N}(0, \bm{\Lambda}^{-1})$, and
let $\bm{Y} \in \mathbb{R}^{n \times p}$ be a data matrix. The posterior
distribution of $\bm{\Lambda}$ is $W_G( b + n,\bm{D} + \bm{Y}^T\bm{Y} )$.

\subsection{Hamiltonian Monte Carlo}
\label{sec:hmc}

Hamiltonian Monte Carlo (HMC) (see, for example, \citep{neal_2010_mcmc}) is an
augmented-variable, MCMC approach to sampling. It can often result in rapid
mixing in distributions with strong correlations between the random variables.
Let $\bm{x} \in \mathbb{R}^{d}$ be a random vector with density $p(\bm{x}) =
Z^{-1} \exp( -E(\bm{x}) )$, where $Z$ is a normalisation constant and
$E(\bm{x})$ is the energy function. By physical analogy, one introduces the
momentum vector $\bm{p} \in \mathbb{R}^{d}$ and constructs the Hamiltonian
function
\begin{equation}
 H(\bm{x},\bm{p}) = E(\bm{x}) + \frac{1}{2} \bm{p}^T \bm{M}^{-1} \bm{p} ,
\end{equation}
where $\bm{M}$ is called the mass matrix. This is to be interpreted as the sum
of a potential energy and a kinetic energy. By sampling from the joint density
$p(\bm{x},\bm{p}) \propto \exp( -H(\bm{x},\bm{p}) )$ and discarding the momenta,
one obtains samples from $p(\bm{x})$.

The procedure for updating the Markov chain is as follows. First, $\bm{p}$ is
sampled from $\mathcal{N}(\bm{0}, \bm{M})$. Then, a new value of $\bm{x}$ is
proposed by simulating Hamiltonian dynamics, as defined by the equations
\begin{equation}
  \dot{\bm{x}} = \bm{M}^{-1}\bm{p} ,
\ \ \ \dot{\bm{p}} = -\nabla E(\bm{x}) .
\end{equation}
The leapfrog integrator is typically used to approximate the dynamics. Given a
step size $\epsilon$ and current state
$\left(\bm{x}_0,\bm{p}_0\right) = \left(\bm{x}(t),\bm{p}(t)\right)$,
the leapfrog algorithm generates a proposal
$\left(\bm{x}_1,\bm{p}_1\right) =
\left(\bm{x}(t+L\epsilon),\bm{p}(t+L\epsilon)\right)$
by iterating the following updates for $L$ steps:
\begin{align}
 \bm{p}\left( \tau+\frac{\epsilon}{2} \right) &=
  \bm{p}(\tau) - \frac{\epsilon}{2} \nabla E\left( \bm{x}(\tau) \right) , \\
 \bm{x}(\tau+\epsilon) &= \bm{x}(\tau) +
  \epsilon \bm{M}^{-1}\bm{p}\left( \tau+\frac{\epsilon}{2} \right) , \\
 \bm{p}(\tau+\epsilon) &= \bm{p}\left( \tau+\frac{\epsilon}{2} \right)
  - \frac{\epsilon}{2} \nabla E\left( \bm{x}(\tau+\epsilon) \right) .
\end{align}
The parameters $(\epsilon,L)$ are chosen manually, typically by performing
preliminary runs.

Finally, the proposal is accepted with probability
$\min\{ 1, \exp( H(\bm{x}_0,\bm{p}_0) - H(\bm{x}_1,\bm{p}_1) ) \}$,
according to Metropolis-Hastings. This is necessary to correct for errors
introduced by inexact simulation of the Hamiltonian dynamics.

\section{Bayesian Inference in GWishart-Based Models}
\label{sec:gw_inf}

A number of sampling procedures have been invented to do inference in the
GWishart distribution. \citet{wang_2010_simulation} developed a rejection
sampler; \citet{mitsakakis_2011_metropolis} developed a Metropolis-Hastings
method in which the proposals are independent of previous samples;
\citet{dobra_2011_bayesian} demonstrated efficiency gains using a random-walk
MCMC scheme in which each element of $\bm{\Lambda}^{\mathcal{V}}$ is updated
conditioned on the other elements.

However, \citet{wang_2012_efficient} demonstrated that a block Gibbs sampler
\citep{piccioni_2000_independence} convincingly outperforms other methods. We
now describe the block Gibbs sampler, and suggest a simple modification to
improve efficiency in high dimensions. We then introduce a novel HMC sampler.
In the final part of this section, we describe a Bayesian sparse GGM in which
these improvements can be utilised.

\subsection{Sampling the GWishart with Block Gibbs}
\label{sec:bg}

The GWishart block Gibbs sampler
\citep{piccioni_2000_independence,wang_2012_efficient}
is based on the fact that a block of $\bm{\Lambda}$ corresponding to a clique
in $G$ can be sampled conditional on the rest of $\bm{\Lambda}$ by sampling
from a Wishart. Thus, given a set of cliques that cover
$\bm{\Lambda}^{\mathcal{V}}$, a sampler for the GWishart can be constructed by
iterating over the covering set and conditionally sampling each block of
$\bm{\Lambda}$.

More precisely, the block Gibbs algorithm is as follows. First, construct a set
$\mathcal{I} = \{\mathcal{I}_k : 1 \leq k \leq K\}$ where $\mathcal{I}_k \in V$
such that all $\mathcal{I}_k$ are cliques, and
$\cup_{\mathcal{I}_k \in \mathcal{I}}
 \bm{\Lambda}_{\mathcal{I}_k,\mathcal{I}_k} = \bm{\Lambda}^{\mathcal{V}}$.
Then, generate the next sample in the Markov chain by iterating over $k$, and
at each step draw
$A \sim W\left( b, \bm{D}_{\mathcal{I}_k,\mathcal{I}_k} \right)$,
and set
\begin{equation}
\bm{\Lambda}_{\mathcal{I}_k,\mathcal{I}_k} = 
 A + \bm{\Lambda}_{\mathcal{I}_k, V \setminus \mathcal{I}_k}
     \left(
      \bm{\Lambda}_{V \setminus \mathcal{I}_k, V \setminus \mathcal{I}_k}
     \right)^{-1}
     \bm{\Lambda}_{V \setminus \mathcal{I}_k, \mathcal{I}_k} .
\end{equation}

The choice of covering set $\mathcal{I}$ can have a significant effect on the
performance of the sampler, and the optimal choice probably depends on $G$.
\citet{wang_2012_efficient} considered two choices for $\mathcal{I}$:
(1) The maximal cliques;
(2) All pairs of nodes connected by an edge, plus all isolated nodes.
It was found that maximal cliques gave better performance than the edgewise
covering set in all the models considered. However, it seems likely that the
set of all maximal cliques will be a suboptimal choice in many models. First,
finding the maximal cliques is NP-hard, so this method may scale poorly.
Second, the number of maximal cliques may be much greater than that required to
cover $\bm{\Lambda}^{\mathcal{V}}$; a smaller covering set may trade off a
little mixing quality for a significant speed up.

In this paper, we investigate this with a simple heuristic to build
$\mathcal{I}$. It is motivated by the desire to build large cliques to
facilitate mixing, but to keep the number of cliques small to reduce run time.
The algorithm is as follows. First, randomly permute the nodes $V$, and permute
the matrix $\bm{G}$ accordingly. Then, iterate over the entries of $\bm{G}$;
for each non-zero $G_{ij}$ not yet included in any clique $\mathcal{I}_k$,
build a maximal clique starting with $\{i,j\}$. Grow this clique by considering
nodes in the permuted order, and add them greedily. So, rather than finding all
maximal cliques, this algorithm quickly finds a small covering set of maximal
cliques, which should improve the efficiency of the block Gibbs sampler in high
dimensional models.

\subsection{Sampling the GWishart with HMC}
\label{sec:gw_hmc}

We develop an HMC method to sample from the GWishart distribution. We now
describe the key choices and challenges to make it work well: dealing with the
positive-definite constraint on the precision, choosing the step size and path
length parameters, and adapting the mass matrix.

\subsubsection{The Positive-Definite Constraint}

The GWishart $W_G(b,\bm{D})$ is a distribution over positive-definite matrices
$\bm{\Lambda} \in M^{+}(G)$. When running HMC, we must ensure that
$\bm{\Lambda}$ remains within the positive-definite cone $S_{++}$. Naively,
this may be done by simply setting the energy to infinity outside $S_{++}$.
However, this may lead to a high rejection rate if the approximate dynamics
often lead to proposals outside the cone. When $b > 2$, the energy approaches
infinity at the boundary, so the chain would remain in $S_{++}$ if the dynamics
could be simulated exactly. But in practice, the leapfrog method can overshoot
the boundary. For $b \leq 2$, the situation is worse as even the exact dynamics
lead the chain outside $S_{++}$.

One idea is to reflect the simulated path off the constraint boundary, which
is straightforward when the constraints are simply bounds on each variable.
But for the positive-definite constraint, it is non-trivial to find where the
path crosses the boundary, or to find the tangent plane at that point. Another
idea is to use the linearly transformed Cholesky
decomposition \citep{atay-kayis_2005_monte}
\begin{equation}
    \bm{D}^{-1}  = \bm{T}^T \bm{T} ;
\ \ \bm{\Lambda} = \bm{\Phi}^T \bm{\Phi} ;
\ \ \bm{\Psi}    = \bm{\Phi} \bm{T}^{-1} ,
\end{equation}
in which the free variables are $\bm{\Psi}^{\mathcal{V}}$.
The non-free variables $\bm{\Psi}^{\bar{\mathcal{V}}}$ are functions of the
free elements that precede them in a row-wise ordering. If HMC is applied in
this representation, there are no constraints on $\bm{\Psi}^{\mathcal{V}}$.
However, the energy gradient with respect to each free
$\Psi_{ij}^{\mathcal{V}}$ depends on the gradients preceding it in the row-wise
ordering. So the gradients must be evaluated sequentially, which makes HMC very
slow.

\newpage
{\bf Step Size and Trajectory Length}

Instead, we run HMC in $\bm{\Lambda}$-space. We find that judicious choices of
the step size and trajectory length are sufficient to acheive good performance.
We draw $\epsilon$ from a distribution that concentrates much of its mass
near the mean, yet still results in the occasional draw of a very small value.
The intuition is that this should facilitate good mixing, while still allowing
the chain to move away from the $S_{++}$ boundary by using a small step size
(and therefore simulating the dynamics accurately). We choose a fixed target
trajectory length $L = \max\left( 1, \lfloor \beta / \epsilon \rceil \right)$,
where $\beta$ is a user-defined parameter, so that when a small $\epsilon$ is
chosen, the chain still moves a long distance. With a fixed trajectory length,
the distribution of $\epsilon$ cannot have too much mass near zero, or the
sampler will be slow. This rules out the exponential distribution, for example.
In this paper, we use a $\Gamma(2,\alpha)$ distribution. The parameters
$(\alpha,\beta)$ are chosen by performing preliminary runs (as is typical in
HMC).

We find this method, in combination with a well-chosen mass matrix, performs
well provided the degrees of freedom parameter $b$ is greater than 10 or so.
This is almost always the case for a posterior GWishart because $b$ increases
with the number of data points. But this HMC approach may not be the best
choice for GWishart priors with small $b$.

\subsubsection{The Mass Matrix}
\label{sec:mass_theory}

To sample from the GWishart distribution, we run HMC on the random vector
$\bm{\Lambda}_{\mathcal{V}}$. We find that the choice of mass strongly
influences its performance. One approach to selecting a mass matrix is to
estimate the covariance $\bm{\Sigma}$ of this random vector and set
$\bm{M} = \bm{\Sigma}^{-1}$ \citep{neal_2010_mcmc}. We considered the following
methods for estimating $\bm{\Sigma}^{-1}$, which we compare experimentally in
section \ref{sec:mass_exp}.
\begin{enumerate}
 \item Set $\bm{\Sigma}$ to the identity matrix.
 \item Perform a short block Gibbs run and set $\bm{\Sigma}$ to the empirical
covariance.
 \item Compute a Laplace approximation to the distribution of
$\bm{\Lambda}_{\mathcal{V}}$, and set $\bm{\Sigma}$ to its covariance.
 \item Assume $\bm{\Lambda} \sim W(b,\bm{D})$, draw samples from the Wishart,
and compute the empirical precision $\bm{K}$ of $\bm{\Lambda}_{\mathcal{W}}$.
Then set $\bm{\Sigma}^{-1} = \bm{K}_{\mathcal{V},\mathcal{V}}$. This
corresponds to a Gaussian approximation on the distribution of
$\bm{\Lambda}_{\mathcal{W}}$ - an approximation which was shown to become more
accurate as $b$ increases \citep{lenkoski_2011_computational}.
\end{enumerate}

\subsection{A Sparse Bayesian GGM}
\label{sec:bayes_ggm}

Consider the following spike-and-slab GGM:
\begin{equation}
 \bm{G} \sim P(\bm{G}) ;
 \ \ \bm{\Lambda} \sim W_G(b,\bm{D}) ;
 \ \ \bm{y} \sim \mathcal{N}(0, \bm{\Lambda}^{-1}) .
\end{equation}
$P(\bm{G})$ is an arbitrary distribution on graphs, but is typically something
simple such as the uniform distribution or a product of Bernoulli distributions
on the edges. This model has received considerable attention in recent years;
see, for example
\citep{dobra_2011_bayesian,rodriguez_2011_sparse,dobra_2011_copula,
wang_2012_efficient}.

Bayesian inference in this model typically involves sampling. One of the most
efficient of such existing methods \citep{wang_2012_efficient}, which we refer
to as WL, iterates two steps: the first samples $\bm{\Lambda}$ given $\bm{G}$
and $\bm{Y}$, and the second makes changes to the graph structure. The first
step is to sample from a GWishart. In the second step, sampling $\bm{G}$ is
done using a double Metropolis-Hastings method \citep{liang_2010_double} with a
reversible jump step. \citet{wang_2012_efficient} propose to flip one edge at a
time, then resample the precision from its prior, before accepting or
rejecting the transition according to Metropolis-Hastings. The efficiency of WL
depends strongly on the efficiency of the GWishart sampler. In this paper, we
improve on existing methods using our HMC technique.

In section \ref{sec:gw_vs_glasso}, we compare the performance of this model
against graphical lasso on a real-world data set, where each model is allocated
the same time budget.

\section{Experiments}

\subsection{HMC versus Block Gibbs}
\label{sec:hmc_vs_bg}

We compare HMC and block Gibbs on synthetically generated data, testing the
effects of dimensionality, data size, and sparsity on the efficiency of these
samplers. Each test case corresponds to a setting of the model dimensionality
$p$; a sparsity parameter $s$ where $0 \leq s \leq 1$; and the ratio $n/q$,
where $n$ is the number of data points and $q$ is the expected number of free
variables.

Each test case is composed of 10 runs. In each run, a graph $G$ is drawn by
sampling each edge from $Bern(s)$, and a precision matrix $\bm{\Lambda}$ is
drawn from $W_G\left(1,p\bm{I}_p\right)$ by taking the $1000^{th}$ sample from a
block Gibbs run. The $n$ data points are then drawn from
$\mathcal{N}\left(\bm{0}, \bm{\Lambda}^{-1}\right)$.

We sample from the posterior of each test run using HMC in which the mass
matrix is computed by sampling from a (fully connected) Wishart distribution
and then conditioning on the missing edges as described in section
\ref{sec:mass_theory}. We compare this to the block Gibbs sampler in which the
covering set of cliques is generated in two different ways:
(1) BG-MC: the covering set consists of all maximal cliques;
(2) BG-HCC: our heuristic clique cover algorithm (see section \ref{sec:bg}).

For all samplers, $\bm{\Lambda}$ is initialised to the identity matrix. We ran
100 iterations of burn-in, and then gathered the following 10000 samples.
Table \ref{tab:syn} shows the results.
\begin{table}[t]
\small
\caption{Comparison by ESS/sec on synthetic data of HMC and two block Gibbs
methods: BG-MC, where the covering set consists of all maximal cliques; and
BG-HCC, our heuristic clique cover algorithm. Block Gibbs is best only for very
low-dimensional models. HMC is orders of magnitude faster in higher
dimensions.}
\label{tab:syn}
\begin{center}
\begin{tabular}{llllllll}
\multicolumn{2}{c}{\bf Test} &
\multicolumn{2}{c}{\bf BG-MC} &
\multicolumn{2}{c}{\bf BG-HCC} &
\multicolumn{2}{c}{\bf HMC}
\\ \hline \\
{\bf Dimension} &
$p$ & ESS & ESS/sec & ESS & ESS/sec & ESS & ESS/sec \\
& 10
 & 7697 (1161) & {\bf 878 (136)}
 & 7086 (1173) & 529 (100)
 & {\bf 10000 (0)} & 514 (10) \\
$n/q = 5$ & 25
 & 4087 (1051) & 39.9 (12.3)
 & 1525 (625)  & 20.0 (8.2)
 & {\bf 10000 (0)} & {\bf 244 (4)} \\
$s = 0.5$ & 50
 & 3037 (693) & 1.59 (0.5)
 & 521 (240)  & 1.26 (0.58)
 & {\bf 9977 (74)} & {\bf 61.8 (3.6)} \\
& 75
 & - & -
 & 188 (81) & 0.151 (0.068)
 & {\bf 9999 (1)} & {\bf 16.6 (0.6)} \\
& 100
 & - & -
 & 150 (76)   & 0.0624 (0.0360)
 & {\bf 8836 (317)} & {\bf 3.62 (0.18)} \\
\\
{\bf Data size} &
$n/q$ & ESS & ESS/sec & ESS & ESS/sec & ESS & ESS/sec \\
& 0.2
 & 4435 (439) & 26.3 (4.1)
 & 1910 (425) & 24.8 (5.3)
 & {\bf 7704 (431)} & {\bf 83.9 (5.5)} \\
$p = 25$ & 1
 & 4060 (1158) & 24.6 (8.4)
 & 1905 (877)  & 26.2 (14.8)
 & {\bf 9990 (16)} & {\bf 235 (34)} \\
$s = 0.5$ & 5
 & 3809 (1084) & 23.1 (7.2)
 & 1681 (640)  & 23.2 (9.9)
 & {\bf 10000 (0)} & {\bf 295 (4)} \\
& 25
 & 3534 (733) & 20.8 (4.6)
 & 1254 (453) & 16.8 (6.2)
 & {\bf 9929 (212)} & {\bf 324 (10)} \\
& 100
 & 3020 (898)  & 18.8 (5.9)
 & 1185 (601)  & 15.7 (8.4)
 & {\bf 9554 (1128)} & {\bf 314 (38)} \\
\\
{\bf Sparsity} &
$s$ & ESS & ESS/sec & ESS & ESS/sec & ESS & ESS/sec \\
& 0.1
 & 8221 (1376) & 147 (34)
 & 8221 (1431) & 146 (35)
 & {\bf 10000 (0)} & {\bf 316 (19)} \\
$p = 25$ & 0.25
 & 3177 (767) & 33.7 (8.7)
 & 2700 (763) & 32.2 (9.5)
 & {\bf 10000 (0)} & {\bf 254 (6)} \\
$n/q = 5$ & 0.5
 & 3089 (1423) & 14.4 (6.8)
 & 1108 (842)  & 12.6 (10.2)
 & {\bf 9995 (13)} & {\bf 239 (7)} \\
& 0.75
 & 7439 (776) & 14.2 (1.7)
 & 1933 (450) & 27.1 (5.2)
 & {\bf 10000 (0)} & {\bf 214 (6)} \\
& 0.9
 & 9426 (179)  & 21.1 (12.7)
 & 4995 (1221) & 96.3 (34.4)
 & {\bf 10000 (0)} & {\bf 205 (21)} \\
\end{tabular}
\end{center}
\end{table}
Times do not include the time taken to compute the index sets in block Gibbs,
or to compute the mass matrix in HMC. When $\bm{G}$ is fixed as in this
experiment, these times are negligible. (But if the sampler is to be part of a
joint sampler for $(\bm{G},\bm{\Lambda})$, then they are significant, as
discussed in sections \ref{sec:mass_exp} and \ref{sec:gw_vs_glasso}).

There are missing entries for BG-MC at dimensionalities 75 and 100: we
abandoned those tests because they were taking an extremely long time. We
consider BG-MC to be impractical for high-dimensional problems. BG-HCC is more
efficient than BG-MC in this scenario, and also when the sparsity is such that
BG-MC has to work with a large number of maximal cliques.

The table shows that BG-MC is best for low dimensional models, but when
$p \geq 25$, HMC is significantly more efficient. The effect of the data size
is different for block Gibbs and HMC. Block Gibbs tends to improve as data size
decreases; HMC tends to improve with more data, we expect because the Gaussian
approximation used when computing the mass matrix becomes more accurate. As the
graph becomes sparser ($s$ decreases), HMC improves because there are fewer
variables to simulate. The block Gibbs methods tend to prefer either very
sparse or very dense graphs, which is expected because these cases will usually
produce fewer cliques than a moderate level of sparsity.

In summary: (1) For block Gibbs, using BG-HCC is preferable to BG-MC in high
dimensions, or when the level of sparsity is unfavourable to BG-MC; (2) Except
for the 10-dimensional case, HMC performs significantly better than both block
Gibbs methods across all our tests.

\subsection{Comparing Methods of Computing the Mass Matrix}
\label{sec:mass_exp}

We compared the methods of computing the mass matrix described in section
\ref{sec:mass_theory}. We generated 10 runs of a single test case as in section
\ref{sec:hmc_vs_bg}. The parameters were: $p=25, n/q=5, s=0.5$. We sampled the
distributions using each of the mass matrix methods. For the two methods
requiring preliminary samples, we drew 20000 points. For the Laplace
approximation, we found the mode numerically by gradient ascent.

The results are shown in Table \ref{tab:mass}.
\begin{table}[t]
\caption{Comparison of methods for computing the HMC mass matrix.}
\label{tab:mass}
\begin{center}
\begin{tabular}{lcccc}
 & {\bf Identity}
 & {\bf Prelim. GWishart}
 & {\bf Laplace}
 & {\bf Prelim. Wishart}
\\ \hline \\
Time to compute M (s) & 0 (0) & 91.6 (13.3) & 27.1 (23.1) & 2.15 (0.04) \\
Sampling time (s) & 5560 (316) & 21.7 (0.2) & 18.3 (0.8) & 21.8 (0.3) \\
ESS      & 2348 (726)    & 10000 (0) & 669 (470)   & 10000 (0) \\
ESS/sec    & 0.425 (0.138) & 461 (3)   & 37.0 (25.9) & 460 (7)   \\
\end{tabular}
\end{center}
\end{table}
Clearly, the identity matrix mass makes HMC highly inefficient. The Laplace
approximation also performed quite poorly. In terms of ESS/sec, a preliminary
sampling run from the GWishart, and from the Wishart (followed by conditioning
on missing edges), gave similar results. However, the preliminary run is
considerably more expensive on the GWishart. If the HMC sampler is to be
embedded in a joint sampler for $(\bm{G},\bm{\Lambda})$, the preliminary
GWishart run needs to be repeated each time the graph changes, which is clearly
impractical. But the precision $\bm{K}$ computed from the Wishart samples
remains valid as the graph changes: the new mass can be obtained simply by
removing those rows and columns from $\bm{K}$ that correspond to missing edges.

\subsection{The Bayesian Sparse GGM versus Graphical Lasso}
\label{sec:gw_vs_glasso}

Having demonstrated the advantages of using an HMC sampler over block Gibbs for
the GWishart, we now employ this method in the Bayesian sparse GGM model, and
compare this model with the frequentist graphical lasso on a real-world data
set. Our data consist of the daily closing prices of 35 stocks from 7 market
sectors of the FTSE100 index on 1000 days over the period from 2005 to 2009.
The price data were cleaned and converted to returns by computing the ratio of
closing prices on consecutive days. The first 500 points were used for
training, leaving 500 points in the test set. We preprocessed the data by
subtracting the mean, and scaling such that the empirical precision had all
ones along the main diagonal.

It is well known that equity returns are not Gaussian-distributed. But for
purposes of comparison, we applied both the Bayesian GGM methods and the
graphical lasso to this data set. To set the penalty parameter $\gamma$ for
graphical lasso, we did 5-fold cross-validation over 100 equally spaced values.
We trained a model with the best-performing $\gamma$ and evaluated it via log
likelihood of the test data.

For the Bayesian GGM, we use the WL method (see section \ref{sec:bayes_ggm}) to
sample $(\bm{G},\bm{\Lambda})$ jointly. We set
$p(\bm{G}|s) = \prod_{i < j} s^{G_{ij}} (1-s)^{(1-G_{ij})}$,
and introduce a sampling step to sample the hyperparameter $s$ (which is
analogous to the L1 penalty in the graphical lasso). We found that the choice
of parameters for the GWishart prior had little effect on the generalisation
performance of the model. We report results for the prior
$W_G(1+n_0,(p+n_0)\bm{I}_p)$, where $n_0 = 10$.

The WL sampler requires samples from both the GWishart prior and posterior,
and its performance depends strongly on the efficiency of this component. We
test both BG-HCC and HMC for this task. BG-MC is far too slow to use within the
WL sampler: when we tried it, WL could not complete a single iteration in the
time it took to complete cross-validation on the graphical lasso. Most of the
time is spent recomputing the maximal cliques each time the graph changes; but
there are many maximal cliques, so lots of time is spent sampling too. For HMC,
our results do not include the time taken to find good settings of the step
parameters $(\alpha,\beta)$. We set these values by adjusting them such that the
acceptance rate on preliminary runs is around 65\% - as suggested by
\citet{neal_2010_mcmc} - and such that the ESS is high. In practice, little
adjustment is needed because our choice of mass matrix means that similar step
parameters can be used for a wide variety of GWishart distributions.

\begin{figure}
\centering
\begin{subfigure}[t]{0.45\textwidth}
 \centering
 \includegraphics[width=\textwidth]{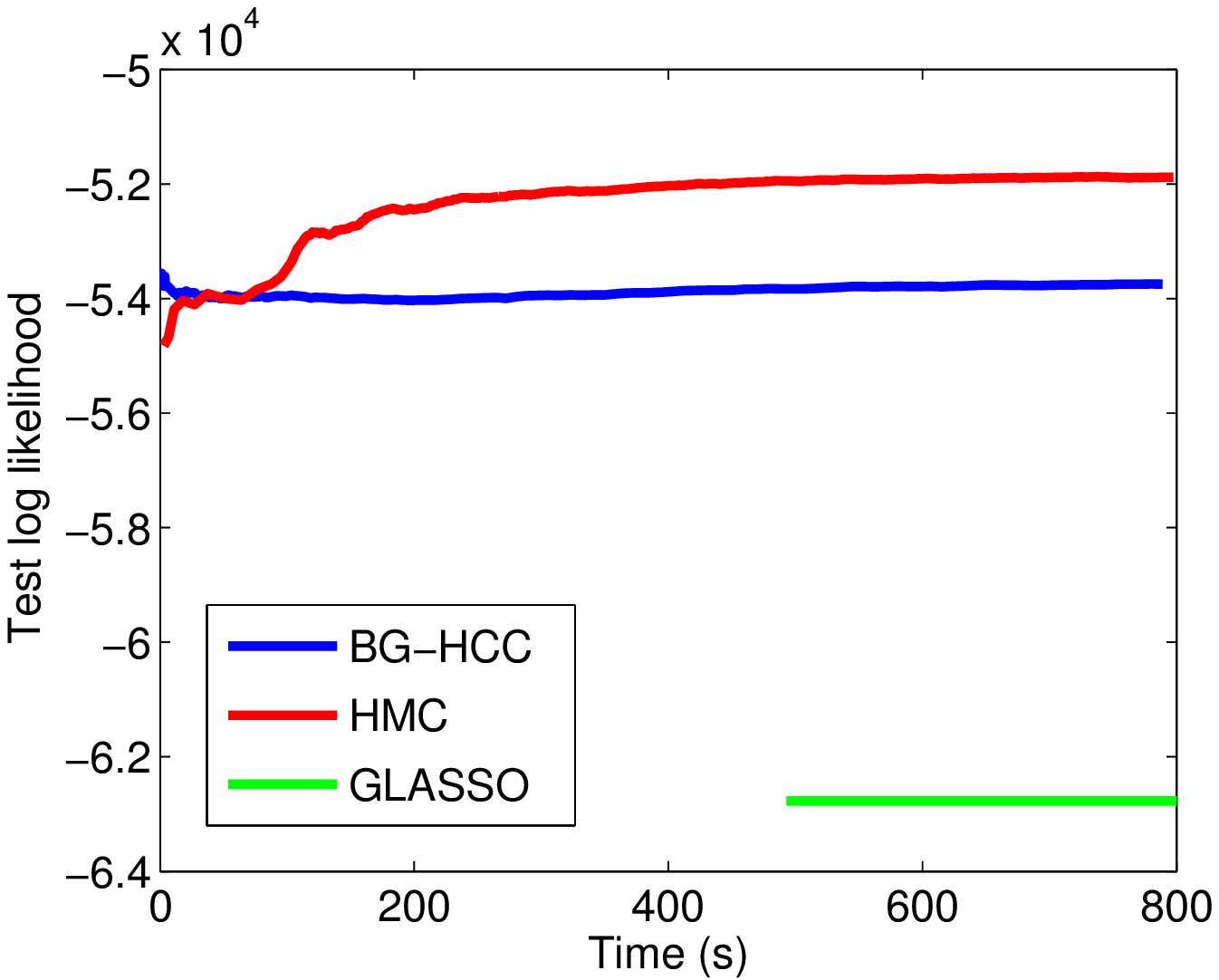}
 \caption{Estimated test log likelihoods over time. The Bayesian methods result
in significantly higher test log likelihoods - good estimates of which are
obtained after only a few samples.}
 \label{fig:llh_vs_time}
\end{subfigure}
\quad
\begin{subfigure}[t]{0.45\textwidth}
 \centering
 \includegraphics[width=\textwidth]{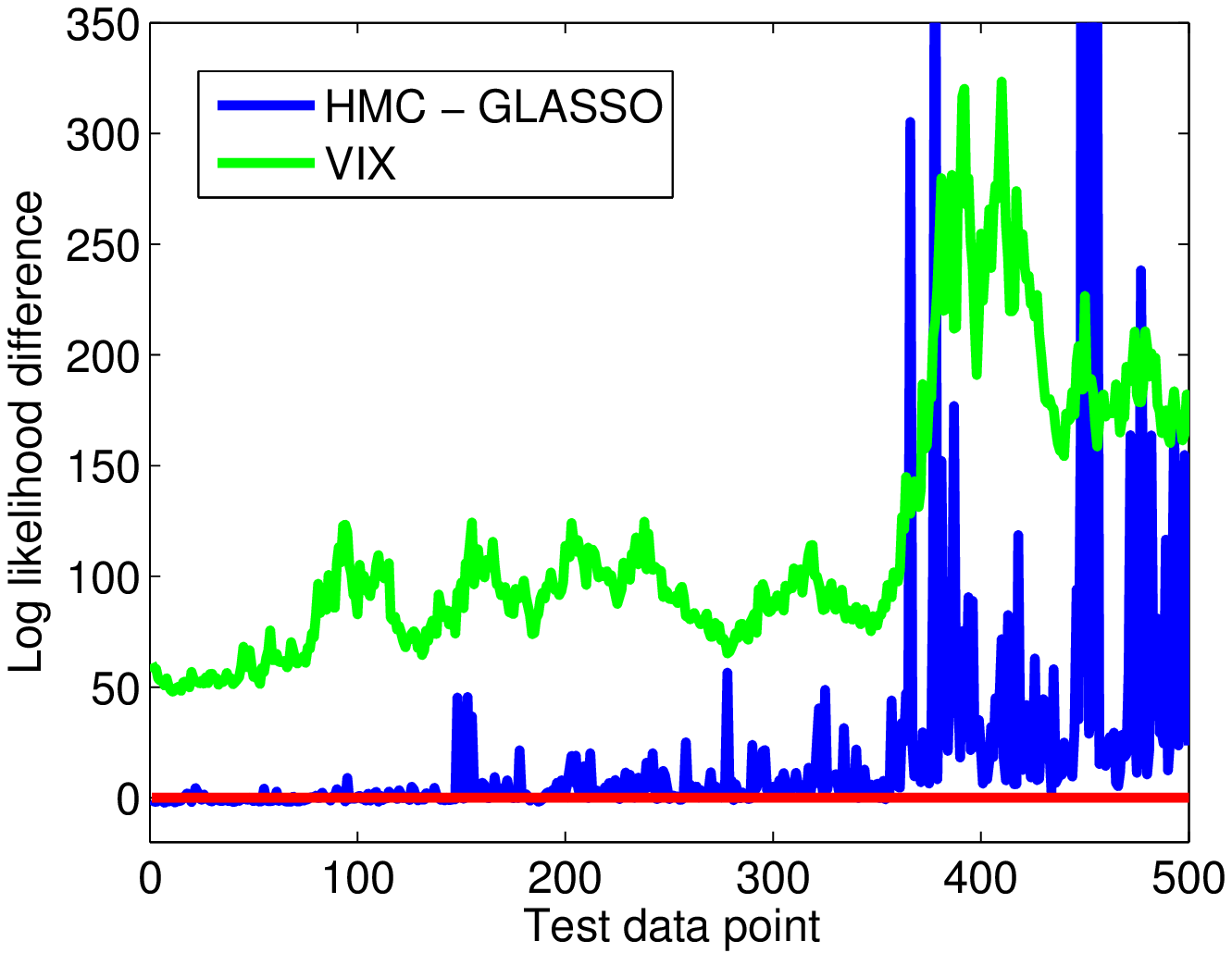}
 \caption{For each point in the test data set, the difference in log likelihood
between the Bayesian GGM and graphical lasso is plotted in blue. The VIX
volatility index, arbitrarily scaled to fit, is in green. The Bayesian method
performs better over the whole test set, but tends to strongly outperform
graphical lasso when market volatility is high.}
 \label{fig:diff_vs_vix}
\end{subfigure}
\caption{
Comparison of the Bayesian GGM and graphical lasso on the FTSE data set.}
\label{fig:ftse_experiment_graphs}
\end{figure}

To make a comparison with graphical lasso, we approximate the expected test log
likelihood
$\langle
  \log p(\bm{Y}_{(test)}|\bm{\Lambda})
 \rangle_{ p(\bm{\Lambda}|\bm{Y}_{(train)}) } \approx
 N^{-1} \sum_{i=1}^N \log p\left( \bm{Y}_{(test)}|\bm{\Lambda}^{(i)} \right)$,
using the samples $\bm{\Lambda}^{(i)}$ drawn from the posterior
$p(\bm{\Lambda}|\bm{Y}_{(train)})$. We compute this expectation after each
sample is drawn; Figure \ref{fig:llh_vs_time} shows how the value evolves over
time for both the HMC and BG-HCC versions of the joint sampler for a typical
run. The samplers do not quite agree because neither has yet converged. But
right from the start, they perform significantly better than graphical lasso,
which takes a few minutes to register its result. At the time graphical lasso
finishes, the test log likelihood scores are:
HMC $= (-5.19 \pm 0.33) \times 10^4$;
BG-HCC $= (-5.38 \pm 0.35) \times 10^4$;
GLASSO $= (-6.28 \pm 0.51) \times 10^4$.
For comparison, the test log likelihood under a Gaussian model with the
empirical precision of the training data is $ (-7.20 \pm 0.70) \times 10^4$.

Figure \ref{fig:diff_vs_vix} offers an explanation for the better performance
of the Bayesian models. It plots the difference in log likelihood for each
point in the test set. The VIX index - which is a measure of market volatility -
is overlayed. The graph shows that the Bayesian GGM fares particularly well
against the graphical lasso when the market is more volatile. This seems likely
to be a result of the Bayesian methods making use of the full posterior, rather
than just using the MAP solution.

\section{Discussion and Future Work}

We have developed a Hamiltonian Monte Carlo sampler for the GWishart
distribution and demonstrated its increased efficiency over the block Gibbs
sampler. Our method - in particular, the choice of mass matrix - is suitable
for embedding into a joint sampler of the graph and precision in a sparse
Gaussian model. We also described a way to choose the covering set in the block
Gibbs sampler that reduced run time and made it more practical to use this
method within a joint sampler.

We then compared a sparse Bayesian GGM model based on the GWishart distribution
with the graphical lasso estimator on a real-world data set. We found that the
Bayesian model performed better in terms of test log likelihood, even when the
models were constrained to the same time budget. The better performance of the
Bayesian model appeared due to its use of the full posterior - as opposed to
the graphical lasso's MAP solution.

Future work could investigate other ways to set the mass matrix, or to select
the step parameters, so as to further improve efficiency. It would also be
interesting to extend the GGM model to incorporate latent variables, analogous
to how \citet{agakov_2012_discriminative} extended the graphical lasso. It
would be harder to choose a useful mass matrix in that case, because the local
geometry would change significantly over the space of joint precision matrices.
Latent variables would also introduce multiple equivalent modes that may be
hard for the sampler to traverse. Another possible direction is to allow the
graph structure or precision matrix distributions to depend on side information.
For example, in the experiment of section \ref{sec:gw_vs_glasso}, market
volatility may influence the dependence structure between the equities. This
could be modelled by allowing the VIX to influence the graph distribution.

\subsubsection*{Acknowledgments}

Peter Orchard wishes to thank Yichuan Zhang and Iain Murray for helpful
discussions - especially those long discussions with Yichuan regarding HMC.

\newpage

\def\bibfont{\small}
\bibliographystyle{apalike}
\bibliography{citations}

\end{document}